\documentclass{article}

\PassOptionsToPackage{numbers,compress}{natbib}
\usepackage[preprint]{neurips_2026}

\usepackage[utf8]{inputenc}
\usepackage[T1]{fontenc}
\usepackage[hidelinks]{hyperref}
\usepackage{url}
\usepackage{booktabs}
\usepackage{amsfonts}
\usepackage{microtype}
\usepackage{xcolor}
\usepackage{subcaption}
\usepackage{graphicx}

\title{Iterate on an LLM, Serve on an SLM}

\author{%
  Keegan Wang\\
  Carnegie Mellon University \quad $\cdot$ \quad Phoebe\\
  \texttt{keeganwa@andrew.cmu.edu}
}

\begin{document}

\maketitle

\begin{center}
\small Accepted to the NeurIPS 2026 2nd Workshop on Agentic AI Benchmarks and Applications for Enterprise Tasks (AABA4ET).
\end{center}

\begin{abstract}
As frontier LLMs gain more capabilities, the cost to run them also goes up. At the same time, small language models (SLMs) also gain more capabilities but have largely been able to keep their cost stable, offering 5-40$\times$ cheaper inference
  for the narrow, repetitive subtasks that agents decompose work into. However, the standing limiting factor is getting the correct behavior out of a small model in a fast, efficient, and economical way. Unfortunately, in most cases, zero-shot small models fail narrow production
  tasks. Furthermore, retraining one every time the behavior changes is too
  slow to iterate on. We propose doing both in sequence. First, develop the desired behavior in prompt space, on a frontier model, with
  immediate feedback from live traffic. Then, once the behavior
  converges, distill the recorded production behavior (not synthetic
  imitations of it) into an SLM through supervised fine-tuning (SFT). We show a proper use case of this method through a scheduling agent for home-care agencies
  that fans out a single yes/no safety decision per caregiver-shift
  pair on GPT-5.4-nano. A LoRA-tuned Qwen3-8B trained on 11.6k
  recorded production decisions reaches 0.913 agreement, while the
  teacher (GPT-5.4-nano), re-run on its own
  recorded prompts, agrees with itself only 72.7\% of the time, pointing to the fact that the SLM may also be more consistent. Furthermore, when
  the two disagreed, a blinded professional scheduler sided with the student
  48-29. Finally, we find that the SLM has ${\sim}2.5\times$ lower median latency and is a fraction of the cost. We believe this lifecycle of quick building,
  live feedback, and cheap inference is a blueprint for others
  building similar systems.
\end{abstract}

\section{Introduction}

Many position papers argue that small language models (SLMs) are the future for agentic systems because agents can decompose
work into narrow, repetitive subtasks or function calls
\citep{belcak2025slm, lu2024slmsurvey} and call cheaper and faster purpose-built models rather than one larger LLM. However, those papers leave an
important question unanswered. How do you get the correct behavior out
of a small model? While a task could be narrow, it doesn't mean that
the task is necessarily easy to develop on. In our production system,
zero-shot models failed our specific task outright (five off-the-shelf SLMs
failed to meet the quality of larger models, and base Qwen3-8B
produced malformed tool calls on 23\% of cases). Furthermore, training a custom
small model for a behavior that you are in the process of designing
requires paying the engineering and training cost again and again
whenever the core behavior changes. If the desired behavior changes mid-training, the process to train an SLM must be restarted. That expense is what makes most
engineers default to a bigger model that just works with minimal
prompting. Yet the savings of SLMs are too large to ignore. With
5-40$\times$ lower cost per call, when any one task or function needs
to be done thousands of times a day, the total cost saved is almost a
no-brainer. So how do we do both?

Our answer is a two-step method to cheaply do both. In phase one, we develop the behavior where iteration
is cheapest: the prompt, skills, and tools. While our task's desired
decision policy was still changing, every prompt change gave us immediate
feedback that we could see live and users could give feedback on. One of the biggest instances was with an early version of our agent in which it approved only 2 of 49 candidates because soft profile tags were acting as vetoes. That same pool moved to 40 after a
one-line prompt change. If we were
re-fine-tuning a model for every change we wanted to make, it would
take days or weeks, not minutes or seconds, to iterate. Then, in phase two, the economics invert. The frontier model is now
an overly large and expensive model, and the small model is the
right-sized, cheaper model to run. At the same time, we have already
recorded tens of thousands of traces and customer behaviors in which we see
approvals and disapprovals in how the agent has acted on its given
task. Importantly, these traces also include many edge cases that would be impossible to have thought about if we were to use only synthetic training data. Now we can distill the behavior of the LLM
into an SLM with data from production rather than synthetic data.

We show this method on a production workload. For each shift a home-care agency is attempting to fill, an agent makes a yes/no decision for each caregiver, up to 1,000 decisions per outreach. After we reach our desired
behavior, we saved 11.6k training examples from over
500 recorded production batches by replaying each prompt, and LoRA-tuned
Qwen3 models on the results. Our fine-tuned Qwen3-8B hit 0.913 agreement
on held-out cases. Interestingly, when we re-ran a random sample of
these traces on GPT-5.4-nano's original prompts, it gave the same
answer only 72.7\% of the time.

Thus, our three contributions are as follows. (1)~A two-phase method where we first develop our desired behavior in prompt space on a frontier model, then distill the production behavior, rather
than synthetic data, into cases to use for fine-tuning. (2)~We re-run the
teacher on its own recorded prompts and use its self-agreement as the
noise floor to decide if the SLM is good enough to use in
production. (3)~A deployed-workload case study with a blinded evaluation by a
professional scheduler that separates quality errors from benign
errors.

\section{The task}

Currently, when a home-care agency wants to fill a client's shift, a human
scheduler must reach out to $N$ caregivers, where $N$ can be in the
hundreds, to ask if they want to take the shift. Rather than have one model reason over hundreds of
candidates with hundreds of lines of information in a single context window, the production system forks one evaluation model per candidate. Each subagent sees the shift's context and request, then has a single tool-call output
deciding to add or reject the caregiver to the shift outreach along with the reasoning.

What makes this decision hard are the
unstructured data, conflicting or missing data, and differing
definitions of ``conflict'' between clients, caregivers,
organizations, and schedulers. Thus we choose to use an LLM rather than a rule-based decision system.

Our production runs this system on GPT-5.4-nano at high reasoning
effort, and each outreach can scale up to 1{,}000 concurrent decisions at a time.

\begin{figure}[t]
  \centering
  \includegraphics[width=0.9\linewidth]{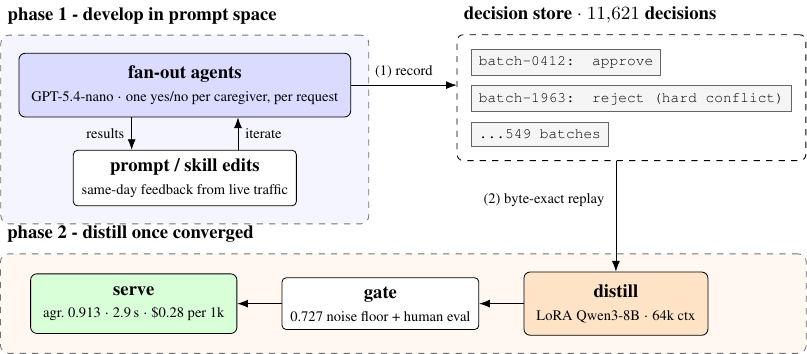}
  \caption{The two-phase lifecycle. Phase 1:
    the deployed agent runs the task on a frontier model while the
    behavior is iterated in prompt space, and
    every trace is recorded. Phase 2: once the behavior converges, the
    stored traces are used as training data,
    and an SLM is distilled from them, gated on the teacher's
    self-agreement noise floor plus a blinded human evaluation.}
  \label{fig:pipeline}
\end{figure}

\section{Training data by byte-exact replay}

To ensure that our training data directly matched production systems,
every production agent decision is durably recorded. Our extractor then replays each
recorded batch through the same production pipeline, using the same prompt builder and serializer that the
live evaluator uses, which produces the identical chat context that
would be served to a deployed model. It then pairs it with the recorded tool
call as the target. This differs from the dominant synthetic-trajectory
approach \citep{wang2023selfinstruct, mukherjee2023orca,
kang2025agentdistill}. Instead, the training distribution is the serving
distribution.

\textbf{Contract filtering.} Recorded rejects in which the reasoning
lacks a hard-conflict justification are purposely dropped to ensure
that training does not teach the violation. We also dedupe decisions
that overlap, keep the most recent decisions, and hold a
train/validation split by source batch. Our training corpus ended up being 549
production batches with 11{,}621 train and 1{,}604 validation
examples across 17 different organizations.

\textbf{Fine-tuning.} We use LoRA \citep{hu2021lora}, at the training
provider's default adapter rank, to fine-tune
Qwen3-4B and Qwen3-8B \citep{yang2025qwen3} on this corpus: constant
LR $10^{-4}$, 1{,}023 optimizer steps ${\approx}$ 1B training tokens
${\approx}$ 3 passes, at 65{,}536 max context. Final eval loss: 0.534
(4B), 0.254 (8B).

\begin{table}[t]
  \caption{Evaluation against recorded production decisions (the 880
    of the 1{,}604 validation cases whose prompts fit the serving
    context, with the teacher noise floor on a seed-fixed random
    150-case subsample of the same set).  $^{*}$Base-model agreement covers only its 676 scored
    cases. The 23\% it failed to answer are excluded, so it reads
    selection-biased high.}
  \label{tab:bakeoff}
  \centering
  \footnotesize
  \setlength{\tabcolsep}{4pt}
  \begin{tabular}{lcccc}
    \toprule
    & Teacher floor & Base 8B & Tuned 4B & Tuned 8B \\
    & (GPT-5.4-nano) & (zero-shot) & & \\
    \midrule
    Agreement w/ recorded decisions & 0.727 & 0.774$^{*}$ & 0.863 & \textbf{0.913} \\
    Malformed outputs & 0 & 203 (23\%) & 106 (12\%) & \textbf{0} \\
    False-approve rate (rec.\ rejects) & 0.218 & 0.208 & 0.231 & 0.232 \\
    False-reject rate (rec.\ approves) & 0.305 & 0.236 & 0.091 & \textbf{0.011} \\
    Hard-conflict prefix compliance & 1.00 & 1.00 & 1.00 & 1.00 \\
    Latency p50/p95 (s) & 7.1 / 13.1 & 14.9 / 36.6 & 2.6 / 3.7 & 2.9 / 4.3 \\
    \bottomrule
  \end{tabular}
\end{table}

\section{Evaluations}

To compare our fine-tuned models to the teacher, and decide
if we could use the new model without regression, we replay held-out
validation prompts with the decision tool forced at temperature 0 and
score against
the held-out production decisions. The live model itself, re-run on a random 150 of its own recorded prompts, only
agreed with itself 72.7\% of the time (95\% CI 0.65-0.79)
\citep{atil2024nondeterminism}. Using this, we can then
interpret our decision agreement, false-approve rate on
recorded rejects, false-reject rate on recorded approves,
hard-conflict prefix compliance, malformed output rate, and latency compared to the teacher.

\section{Results}

We have four main results (Table~\ref{tab:bakeoff}). The teacher (the
production GPT-5.4-nano agent) scores against its own recorded
decisions as a self-agreement noise floor. The tuned 8B Qwen model
scoring 0.913 (95\% CI 0.89-0.93) is more consistent with the existing
traces than the teacher itself. When measured against the teacher's original decisions, the disagreement concentrates in the benign direction (false
rejects) and the false-approve rate is flat across each arm. While
the 4B model had decent decision making, its 12\% malformed
output rate makes it a bad candidate for production use.
Therefore the capability floor likely sits between 4B and 8B parameters.

We also find that the zero-shot, un-fine-tuned Qwen3-8B fails to produce a usable tool call on 23\% of cases and takes
a median of 14.9\,s before outputting. The tuned model eliminates the malformed outputs and answers
5$\times$ faster, so using Qwen3-8B without the fine-tuning would not
have been a viable alternative.

\subsection{Human ground truth}

Next, to test the agents' correctness against the real world, we ran a
blinded 192-case evaluation by a professional home-care scheduler. We included all 77
student-teacher disagreements as well as stratified samples of
agreeing rejects (49) and approves (66) to ground the comparison in human
judgement. Because the sample deliberately oversamples
disagreements, we post-stratify by the student-vs-recorded outcome,
whose population sizes over the full 880-case validation set are
known (77 disagreements, 235 agreeing rejects, 568 agreeing
approves).

\textbf{Head-to-head disagreement.} On the 77 disagreement cases the
human sided with the tuned 8B 48 times vs.\ 29 for the production
decision (62\%-38\%, 95\% CI 0.51-0.72, exact binomial two-sided
$p=0.040$).

\begin{table}[t]
  \caption{Accuracy compared to human labels, stratified and
    weighted to the full 880-case validation set. Error directions also reported for the two arms that decide every case.}
  \label{tab:human}
  \centering
  \small
  \begin{tabular}{lcccc}
    \toprule
    vs.\ human labels & Teacher (recorded) & Base 8B & Tuned 4B & Tuned 8B \\
    \midrule
    Accuracy (reweighted) & 0.866 & 0.705 & 0.797 & \textbf{0.888} \\
    False-approve (human rejects) & \textbf{0.199} & --- & --- & 0.273 \\
    False-reject (human approves) & 0.098 & --- & --- & \textbf{0.024} \\
    \bottomrule
  \end{tabular}
\end{table}

\textbf{What human grounding reveals.} Against the teacher's own
labels, the student's false-approve rate looked flat. However, against
human labels, an asymmetry appears (Table~\ref{tab:human}). The
student approves somewhat more caregivers the human would reject
(0.273 vs.\ 0.199) while cutting false rejects from 0.098 to 0.024.
The net difference is more accurate and far less over-filtering, but
slightly more permissive toward caregivers that a human would not add
to an outreach. Note that we generally find that humans add fewer caregivers to outreaches than LLMs, even if all caregivers are good fits.

\begin{figure}[t]
  \centering
  \begin{subfigure}[t]{0.53\linewidth}
    \centering
    \includegraphics[width=\linewidth]{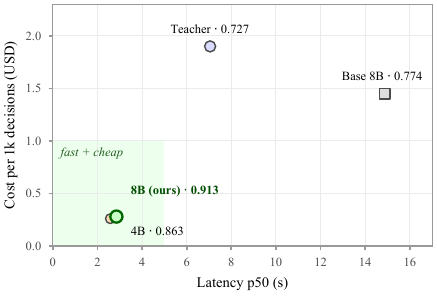}
    \caption{Cost and latency (labels: agreement).}
    \label{fig:econ-cost}
  \end{subfigure}\hfill
  \begin{subfigure}[t]{0.44\linewidth}
    \centering
    \includegraphics[width=\linewidth]{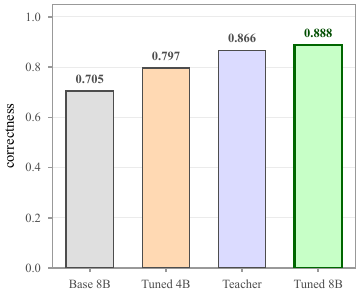}
    \caption{Correctness vs.\ human labels.}
    \label{fig:econ-correct}
  \end{subfigure}
  \caption{The four arms compared. (a)~Only the tuned models land in
    the fast-and-cheap quadrant (shaded). The tuned 8B is
    ${\sim}7\times$ cheaper and $2.5\times$ faster than the teacher.
    Teacher plotted at its cached-prompt cost (\$1.9 per 1k, \$7.7
    uncached). SLM costs are price estimates at full utilization
    based on industry virtual GPU pricing.
    (b)~Accuracy against the blinded human labels, post-stratified and
    reweighted to the full validation set (Table~\ref{tab:human}): the
    tuned 8B is the most correct model, and the
    zero-shot base is the least correct arm.}
  \label{fig:econ}
\end{figure}

\subsection{Cost and latency}

Figure~\ref{fig:econ-cost} places the four arms on the cost-latency
plane. At list prices and measured token distributions, the teacher
costs ${\sim}\$1.9$-$\$7.7$ per 1k decisions depending on
prompt-cache state. A dedicated single-GPU deployment of the tuned 8B
at the observed evaluation throughput lands at
${\sim}\$0.19$-$\$0.38$ per 1k at full utilization (A100- and
H100-class GPU-hour prices). The tuned 8B
is better than every other usable arm. It is simultaneously the cheapest
of the arms that answer every case,
the most consistent (0.913), and within a third of a second of the
fastest arm (the 4B, ruled out by its malformed outputs), while the
zero-shot base model is dominated on every axis.

\section{Related work}

\textbf{SLMs for agents.} \citet{belcak2025slm} argue agentic
workloads are SLM-shaped. Surveys map the capability landscape
\citep{lu2024slmsurvey}. We contribute a deployed instance with
measured stakes. 

\textbf{Distillation from teacher outputs.} Classic
knowledge distillation \citep{hinton2015distilling} and its LLM
lineage (rationale distillation \citep{hsieh2023distilling},
instruction imitation \citep{wang2023selfinstruct, mukherjee2023orca},
and agent-trajectory distillation \citep{kang2025agentdistill,
zeng2023agenttuning}) mostly train on synthetic teacher generations.
Our targets are recorded production decisions replayed byte-exactly.

\textbf{Tool-use tuning of small models} \citep{patil2023gorilla,
qin2023toolllm, liu2024apigen, liu2024toolace, erdogan2024tinyagent}
shows small models can match frontier function calling. TinyAgent
\citep{erdogan2024tinyagent} is the closest deployed example.

\textbf{Students surpassing noisy teachers.} Born-again networks
\citep{furlanello2018born}, noisy student \citep{xie2020noisystudent},
and weak-to-strong generalization \citep{burns2023weaktostrong} all
show that a student can end up better than the teacher it learned
from. We see the same thing in a production setting. Since the teacher
only agrees with itself 72.7\% of the time
\citep{wang2023selfconsistency, zheng2023judging,
atil2024nondeterminism}, it acts less like ground truth and more like
a noisy annotator \citep{dawid1979mle, song2022noisylabels,
plank2022label}.

\textbf{Context distillation} \citep{askell2021lab, snell2022context}
takes what a prompt does and trains it into the model's weights. Our
method does the same thing one level up. Instead of a single prompt,
we distill a whole behavior that was shaped through prompt iteration.

\textbf{Cost-motivated routing} \citep{chen2023frugalgpt,
ong2024routellm} cuts costs by sending easier requests to cheaper
models. We instead replace the entire subtask with one small model.
Industry case studies on compressing models for production serving
report the same kind of savings we find \citep{behdin2025scaling}.

\section{Limitations and conclusion}

\textbf{Limitations.} As of writing this paper, the distilled model is not yet serving
production traffic. Every number here is an offline replay, and the
true quality test is an online A/B test in production, where prompt-cache dynamics, traffic mix, and downstream
scheduler behavior all contribute to cost, correctness, and latency. Our ground-truth
labels come from a single professional scheduler, and our statistics
do not model correlation between cases from the same organization or
batch. Finally,
results cover one subtask, one domain, and one teacher. Thus, further work is needed to prove this method effective for a wide range of engineering tasks.

\textbf{Conclusion.} To make good SLM systems, one should first
develop the behavior where iteration is cheap (in prompt space on a
frontier model) and, once the behavior converges, compile it into a
small model trained on the recorded behavior itself. On our
production workload, this yielded an 8B SLM that executes the policy more
reliably than the teacher that defined it, at a fraction of the
cost and latency, and, by blinded expert judgement, more accurately
as well. This protocol of quick building, live feedback,
then cheap inference should be a blueprint for others building similar
systems.

\bibliographystyle{plainnat}
\bibliography{references}

\end{document}